\documentclass[letterpaper, 10 pt, conference]{ieeeconf}  
\usepackage{microtype}
\usepackage{graphicx}
\usepackage{booktabs}
\usepackage{hyperref}
\usepackage{units}
\usepackage{subcaption}
\usepackage{pgfplots}
\usetikzlibrary{backgrounds, calc}
\pgfplotsset{compat=1.16}

\hypersetup{
    colorlinks=true,
    linkcolor=blue,
    filecolor=magenta,      
    urlcolor=cyan,
}

\usepackage{amsmath}
\usepackage{amssymb}
\usepackage{xcolor}
\usepackage{soul}
\usepackage[nolist,nohyperlinks]{acronym}

\begin{acronym}[fancy]
\acro{fancy}[GS-CIOC]{General-Sum Multi-Agent Continuous Inverse Optimal Control}
\end{acronym}

\usepackage[framemethod=tikz]{mdframed}
\usepackage{makecell}

\usepackage{algorithm, algorithmic}

\definecolor{mycolor}{rgb}{0.122, 0.435, 0.698}

\IEEEoverridecommandlockouts                              

\makeatletter
\def\title@font{\Large\bfseries}
\let\ltx@maketitle\@maketitle
\def\@maketitle{\bgroup
	\let\ltx@title\@title
	\def\@title{\resizebox{\textwidth}{!}{
			\mbox{\title@font\ltx@title}%
	}}%
	\ltx@maketitle
	\egroup}
\makeatother

\title{\LARGE \bf
Real-Time Environment Condition Classification for Autonomous Vehicles}

\author{Marco Introvigne$^{1*}$, Andrea Ramazzina$^{1*}$, Stefanie Walz$^{1*}$, Dominik Scheuble$^{1*}$, Mario Bijelic$^{2*}$
\thanks{$^{*}$All authors contributed equally to this work.}%
\thanks{The authors are with $^{1}$Mercedes-Benz AG and $^{2}$Princeton University.}%
}

\begin{document}

\usepgfplotslibrary{fillbetween}

\maketitle
\thispagestyle{empty}
\pagestyle{empty}

\begin{abstract}
Current autonomous driving technologies are being rolled out in geo-fenced areas with well-defined operation conditions such as time of operation, area, weather conditions and road conditions. In this way, challenging conditions as adverse weather, slippery road or densely-populated city centers can be excluded. In order to lift the geo-fenced restriction and allow a more dynamic availability of autonomous driving functions, it is necessary for the vehicle to autonomously perform an environment condition assessment in real time to identify when the system cannot operate safely and either stop operation or require the resting passenger to take control. 
In particular, adverse-weather challenges are a fundamental limitation as sensor performance degenerates quickly, prohibiting the use of sensors such as cameras to locate and monitor road signs, pedestrians or other vehicles.
To address this issue, we train a deep learning model to identify outdoor weather and dangerous road conditions, enabling a quick reaction to new situations and environments. We achieve this by introducing an improved taxonomy and label hierarchy for a state-of-the-art adverse-weather dataset, relabelling it with a novel semi-automated labeling pipeline. 
Using the novel proposed dataset and hierarchy, we train RECNet, a deep learning model for the classification of environment conditions from a single RGB frame. We outperform baseline models by relative \textbf{16}\% in F1-Score, while maintaining a real-time capable performance of 20 Hz.
The code is published \href{https://github.com/marcointrovigne/WeatherDetection}{here}\footnote{\scriptsize\url{ https://github.com/marcointrovigne/WeatherDetection}\label{link}}.
\end{abstract}

\section{Introduction}
While autonomous driving technology has remarkably improved over the last years, there are many open problems that need to be solved in order to deploy such technology worldwide. Examples of such unsolved challenges are foggy or rainy conditions \cite{volk2019rainObj,hnewa2020rain}, night-time settings \cite{jhong2021nighttime,morawski2021nod} as well as densely-populated areas \cite{wang2018repulsion,xu2023objectCrowd}. To circumvent these current limitations, autonomous driving functionalities are being rolled out in geo-fenced areas. In such settings, it is critical to have a reliable approach to assess the environmental conditions, so to detect if one critical circumstance is met \cite{charmet2023overviewODD,gurumurthy2020geofencing} and shut the operation or require the resting driver to take back control. Furthermore, as such dynamic environment conditions might rapidly change, it is critical for this system to operate in real-time on the vehicle while being computationally lightweight.
In particular, weather as one of those limiting factors significantly impairs the sensor performance in dynamic spray conditions with slippery roads, fog, snow, rain, a glaring sun and nighttime conditions. Estimating those conditions from RGB camera images can support the decision-making, improving the roll-out of autonomous vehicles to broader areas. \newline

\begin{figure}
    \centering
    \includegraphics[width=\linewidth]{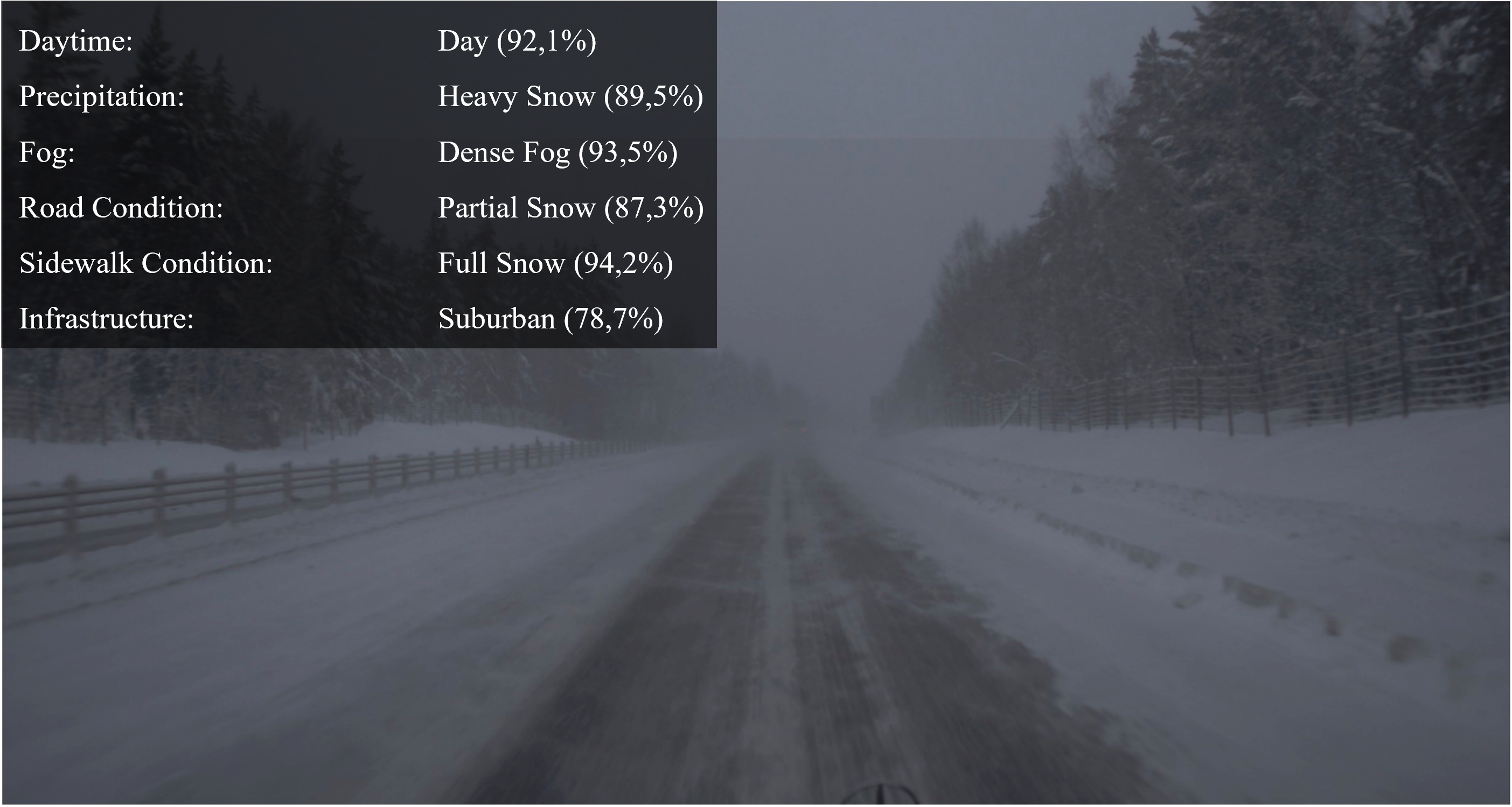}
    \caption{We introduce RECNet, a unified model for real-time classification of environment conditions from a single RGB frame. This is enabled by our novel taxonomy and our proposed semi-automated labeling pipeline.}
    \vspace{-1eM}
    \label{fig:teaser}
\end{figure}

Conventional ways of weather identification mainly relied on several sensors \cite{weathersensors} to account for all the variables simultaneously. However, sensor installation and maintenance require a significant workforce and material resources. As a result, weather detection technologies based on image processing and computer vision have steadily evolved \cite{weathernet,weatherclassification1, weatherclassification2, weatherclassification3}.
However, such approaches are limited in scope \cite{weatherclassification1, weatherclassification3} and cover only a subset of possible conditions encountered in the real-world settings \cite{weathernet,weatherclassification2}. \newline
We solve this limitation by building on top of the DENSE dataset introduced in \cite{SeeThrough}, by improving the label taxonomy and re-annotating the dataset with a semi-automated labelling pipeline incorporating Lidar point cloud data along with RGB images to refine the differentiation of complex weather scenarios. This enables us to train a real-time capable unified model, dubbed RECNet (\textbf{R}ealtime \textbf{E}nvironment \textbf{C}ondition \textbf{Net}work), able to classify the different environmental conditions. Extensive experiments and ablation studies demonstrate the effectiveness of our approach.\newline
In summary, we make the following contributions:
\begin{itemize}
    \item We introduce a novel environment condition dataset, dubbed DENSE++, comprising of six categories: daytime, precipitation, fog, road condition, roadside condition and scene-setting. 
    \item We introduce a semi-automated labeling pipeline complementing the human annotator capabilities, in particular with respect to the estimation of precipitation intensity.
    \item We devise RECNet, a lightweight deep learning model for environment condition classification from a single RGB frame, and we ablate the model architecture and hyper-parameters choices. Our model reaches $92\%$ accuracy, while running at 20fps on a single GPU.  
\end{itemize}

\section{Related Work}\label{sec:relwork}
\subsection{Classification}

For image-based environment condition classification in ADAS functionalities, a backbone for efficient and real-time capable feature extraction is required.
Often, backbones are based around a ResNet architectures \cite{he2016deep} using residual connections between different layers that mitigate the vanishing gradient problem. 
These connections enable the training of deeper networks, fostering more accurate and efficient classifiers, such as those used in weather recognition systems.
Residual connections are often paired with convolution layers to create common building blocks of backbones.
However, when too many of these blocks are concatenated in a deep feature extractor, the number of parameters, thus the computational load increases.
Hence, \cite{mobilenet} introduced MBConv, where the convectional layers are organized such  that the number of parameters is reduced, allowing high computational efficiency and real-time capability while keeping the high representative capability of ordinary convolutional layers with residual connections.
A popular classificator relying on MBConv blocks is EfficientNet \cite{efficientnet}.
They introduce a novel scaling method that optimizes model depth, width, and resolution simultaneously, effectively scaling up the network without significantly increasing computational costs; thus allowing real-time capability for accurate classification results. 

An alternative to backbones relying on convolution layers is Vision Transformers (ViT) \cite{transformergoogle}.
ViT employs the transformer architecture by representing images as sequences of image patches. 
Unlike traditional transformers, ViT addresses computational challenges by using global attention over image patches.
However, the attention mechanism of Vision Transformers is still computationally more expensive then convolutions; hence limiting their real-time applicability on current hardware.

\subsection{Datasets}

A prominent dataset for weather classification is the Image2Weather dataset introduced by Chu~et~al. \cite{image2weather}.
The dataset comprises over 180,000 photos associated with weather information  obtained by linking the image's time and geographical details with data from a weather forecast website.
However, the captured scenarios are often unrelated to autonomous driving scenarios and relying on weather forecast as ground truth is insufficient as the prominent weather condition might change frequently. 
Similar image-based datasets are provided by \cite{guerra2018weather}.

Datasets targeted towards autonomous driving like KITTI \cite{geiger2013vision}, CityScapes \cite{cordts2016cityscapes} or Waymo \cite{sun2020scalability} often only include clear-weather scenarios or are biased towards good weather.
If they do include adverse weather scenarios, like nuScenes \cite{caesar2020nuscenes} or Argoverse \cite{chang2019argoverse}, the provided annotation of the weather is often very coarse and no measure of the intensity of the weather effect is provided.
Datasets that are targeted towards adverse weather conditions, often feature only a limited number of frames, e.g. ACDC \cite{sakaridis2021acdc} constituting of only 4000 frames.
Other datasets like WEDGE \cite{marathe2023wedge} add synthetic adverse weather effects using generative neural networks, but limit themselves to single sensor streams.

Large adverse weather datasets captured in-the-wild are e.g. the CADC dataset \cite{pitropov2021canadian} that is however limited to a single geographic region.
A very prominent dataset in this context is DENSE \cite{SeeThrough}, a multi-modal dataset covering over 10,000 km of driving in northern Europe, providing comprehensive annotations for different weather conditions as well as 3D object annotations.

\subsection{Weather Classification}

The accurate detection of environmental conditions is pivotal in decision-making for autonomous vehicles. However, the existing literature lacks integration for the simultaneous classification of weather, road conditions, and scene settings using a single algorithm.

For weather classification, Ibrahim et al. introduce WeatherNet \cite{weathernet}, a network of four deep Convolutional Neural Network (CNN) models with ResNet50 architecture.
Typically however, traditional image classification methods fall short in weather recognition, where multiple weather features coexist. 
Thus, Li et al. \cite{weatherimagesegmentation} propose a CNN-based multi-task framework for simultaneous weather classification and weather-cues semantic segmentation. 
Another approach by Li et al. \cite{weatherfeatureextraction} involve fusing weather-specific and CNN features for multi-class weather classification.
Zhang et al. \cite{evolalg} introduce an evolutionary algorithm into EfficientNet \cite{efficientnet} for solving the multi-weather classification problem.
Zhao et al. \cite{weatherrnn} address the issue with a CNN-RNN-based multi-label classification approach, identifying relationships between weather classes.
Pan et al. \cite{road1} employ VGG16 \cite{simonyan2014very} to categorize road snow quantity, while Choi et al. \cite{road2} avoid imbalances between different weather effects by using CycleGAN \cite{zhu2017unpaired} to transform dry road surface images into wet and snowy ones, augmenting the dataset at minimal cost. In summary, previous methods fall short in subdividing atmospheric, particle and road effects. With the new introduced labeling hierarchy those effects can be annotated simultaneously and jointly predicted, taking their complementing character into account. 
\section{Method}

The contributions for making RECNet an effective data-driven environment classification method are twofold. First, we introduce a novel weather classification dataset comprising of diverse environment conditions paired with detailed annotations in Sec.~\ref{sec:dataset}.
Second, we present architecture and training details in  Sec.~\ref{sec:classiffication}.
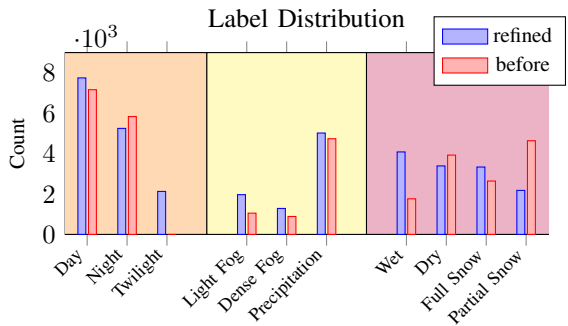
\begin{figure}
    \centering
\vspace{1mm}
\begin{tikzpicture}
\begin{axis}[
    ybar=1pt, 
    bar width=3pt, 
    enlarge x limits=0.05, 
    legend style={at={(0.9,1.2)},
      anchor=north},
    ylabel={Count},
    ymin=0, ymax=9000, 
    symbolic x coords={Pre-Day, Day, Night, Twilight, Pre-LightFog, Light Fog, Dense Fog, Precipitation, Pre-Wet, Wet, Dry, Full Snow, Partial Snow, PostPartialSnow},
    xtick=data,
    x tick label style={rotate=45,anchor=east},
    width=0.45\textwidth, 
    height=4cm,
    area legend, 
    set layers,  
    title={Label Distribution}, 
    xticklabel style={align=center},
    xticklabels={
    \scriptsize{Day}, 
    \scriptsize{Night}, \scriptsize{Twilight}, 
    \scriptsize{Light Fog}, 
    \scriptsize{Dense Fog}, \scriptsize{Precipitation},
    \scriptsize{Wet}, 
    \scriptsize{Dry}, 
    \scriptsize{Full Snow}, 
    \scriptsize{Partial Snow},},
    scaled y ticks=base 10:-3,
    yticklabel style={
      /pgf/number format/fixed,
      /pgf/number format/precision=0,
      /pgf/number format/fixed zerofill,
      /pgf/number format/1000 sep=k 
    },
    label style={font=\footnotesize}
    ]

\draw[fill=orange!30] (axis cs:Pre-Day,\pgfkeysvalueof{/pgfplots/ymin}) rectangle (axis cs:Pre-LightFog,\pgfkeysvalueof{/pgfplots/ymax});

\draw[fill=yellow!30] (axis cs:Pre-LightFog,\pgfkeysvalueof{/pgfplots/ymin}) rectangle (axis cs:Pre-Wet,\pgfkeysvalueof{/pgfplots/ymax});

\draw[fill=purple!30] (axis cs:Pre-Wet,\pgfkeysvalueof{/pgfplots/ymin}) rectangle (axis cs:PostPartialSnow,\pgfkeysvalueof{/pgfplots/ymax});

\addplot+[ybar] plot coordinates {(Day, 7748) (Night, 5249) (Twilight, 2128) (Light Fog, 1968) (Dense Fog, 1286)  (Precipitation, 5019) (Wet, 4083) (Dry, 3390) (Full Snow, 3335) (Partial Snow, 2179)};
\addplot+[ybar] plot coordinates {(Day, 7163) (Night, 5834) (Twilight, 0) (Light Fog, 1052) (Dense Fog, 887) (Precipitation, 4736) (Wet, 1762) (Dry, 3926) (Full Snow, 2642) (Partial Snow, 4637)};

\legend{\footnotesize{refined}, \footnotesize{before}}
\end{axis}
\end{tikzpicture}
\vspace{-2mm}
\caption{Distribution of the refined and previous labels for the overarching categories: daytime, weather, and road conditions. The additional "twilight" tag clearly separates day from night. Precipitation is defined such that it can occur simultaneously with fog, leading to improved scene descriptions. Finally, separating road and sidewalk conditions improves accuracy, particularly for scenarios like dry roads with snow-covered sidewalks, which were previously grouped together.
}\label{fig:class-distribution.}
\end{figure}
\begin{figure}[t!]
    \centering
    \includegraphics[width=\linewidth]{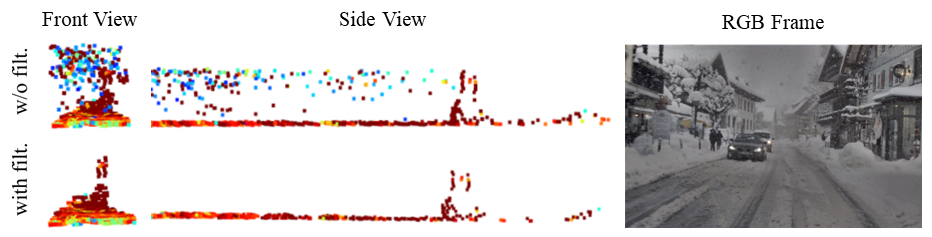}
    \vspace{-4mm}
    \caption{Precipitation Intensity Annotation. Lidar points without a sufficient number of neighborhood points are classified as stemming from precipitation, as visualized for a heavy snow scene (right). Upper and lower row show point clouds without and with clutter points removed.}
    \label{fig:dror}
    \vspace{-2mm}
\end{figure}

\subsection{Dataset}
\label{sec:dataset}
\begin{figure}[!t]
    \centering
    \vspace{1mm}
    \includegraphics[width=\linewidth]{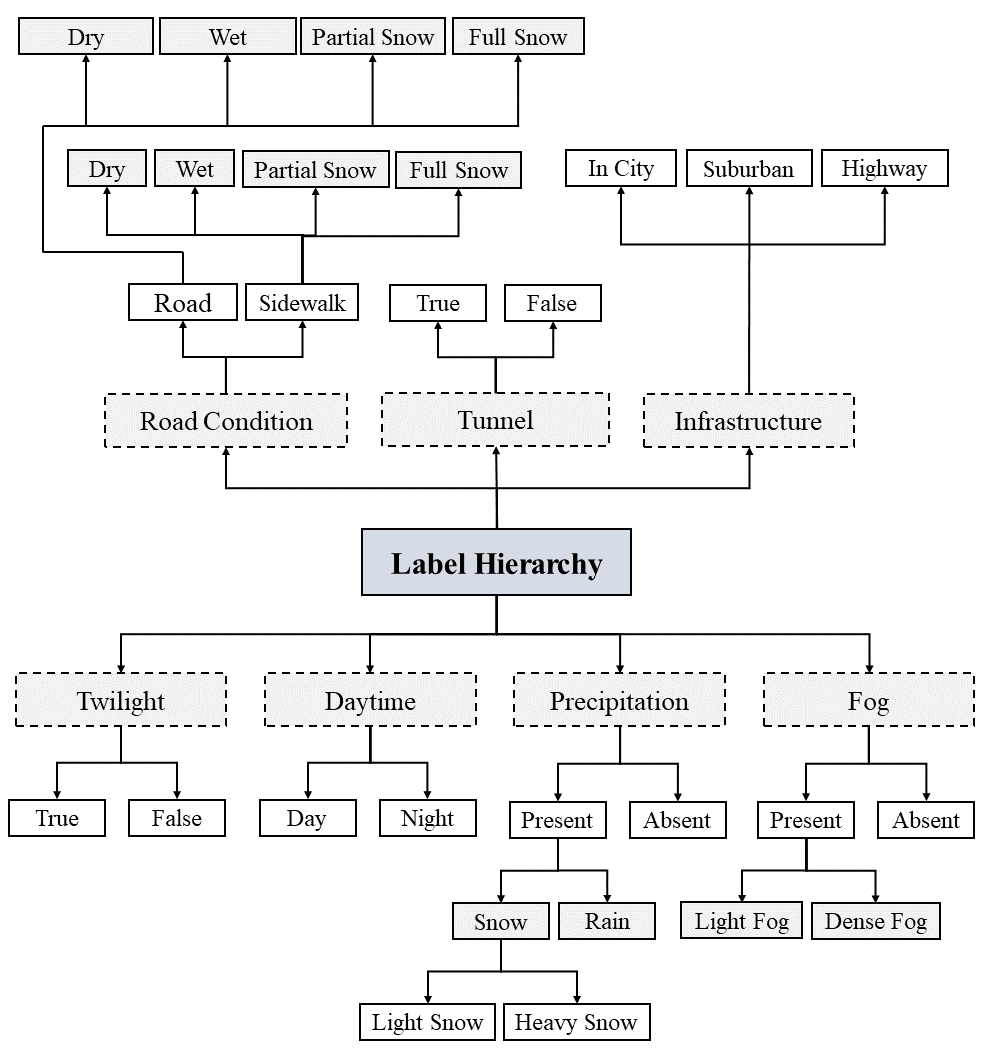}
    \caption{Environment classification hierarchy. The newly introduced label hierarchy allows to assign multiple weather conditions to a single frame, introduces a classification of the precipitation intensity, and differentiates between road and sidewalk.}
    \label{fig:label_hierarchie}
    \vspace{-1mm}
\end{figure}

We build our label hierarchy on top of the popular DENSE dataset \cite{SeeThrough} because of its extensive diversity in environmental conditions and adverse weather effects. Although annotations for these conditions are provided, their level of detail is simple in comparison to the added structure. 
In particular, the previous labeling falls short by not supporting multiple weather effects simultaneously, despite e.g. fog frequently occurring with rain \cite{tardif2008process, ali2004fog}. Additionally, classifying a single global road condition proves to be insufficient; for example, sidewalks may remain snowy or icy even when the roads have been cleared. This lack of clarity in the annotations necessitates a novel, more refined labeling hierarchy presented in Fig.~\ref{fig:label_hierarchie}.
\begin{figure*}[!t]
    \centering
    \includegraphics[width=1\textwidth]{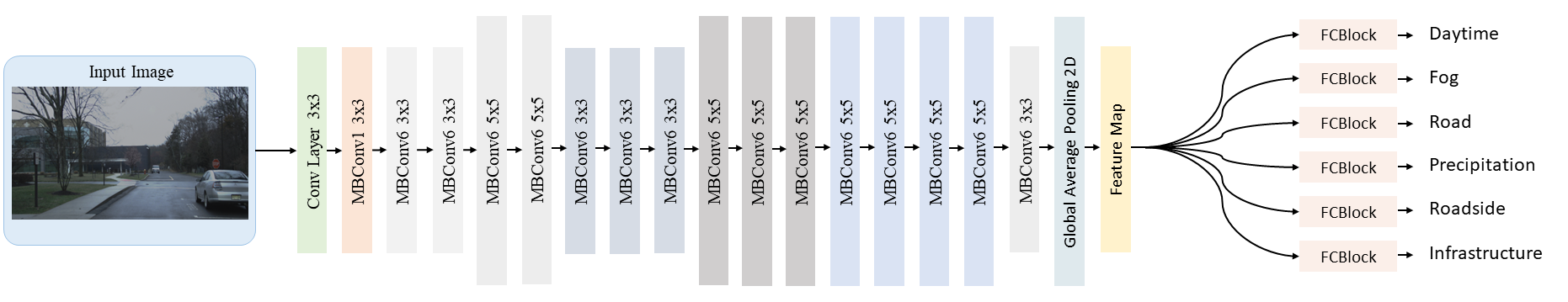}
    \vspace{-3mm}
    \caption{Network overview figure for the proposed RECNet. As the backbone, we use the EfficientNet-B2 network, a lightweight deep learning model capable of extracting a general feature map. This map is utilized by the six downstream classification units to classify all annotated environmental conditions incl. Fog, Daytime, Road Conditions, Precipitation and Infrastructure. }
    \label{fig:Architecture}
\end{figure*}
To this end, we improve the taxonomy of the label hierarchy by first differentiating between fog and precipitation. Perception itself is distinguished into snow and rain that can be each classified into light or dense. Furthermore, we distinguish more finely the road conditions by differentiating between road and sidewalk.

With the new label hierarchy established, we annotated all 12,997 keyframes using manual and automated labeling methods. Human annotators rely on the RGB images and determine the specific road conditions for both the road and sidewalk in each frame. However, distinguishing precipitation intensity from RGB images or other raw sensor data poses a significant challenge for human annotators, necessitating a more data-driven approach. Therefore, we use point clouds from the captured Velodyne HDL-64 to count the backscattered points as a precipitation indicator. In detail, precipitation introduces many clutter points in the point cloud, and the quantity and spatial distribution of these clutter points vary with the type and intensity of precipitation, which we utilize to measure the intensity. To count the clutter points, we adopted the approach of the DROR filter \cite{pointremove}.

Following heuristics, points on objects typically have several other points nearby.
In contrast, points from snowflakes and raindrops do not. To this end, for every point, we count the number of points within a spherical neighborhood defined by the search radius \(r_s\):
\begin{equation}
    r_s = \alpha \beta \frac{\pi r_p}{180},
\end{equation}
where \(\alpha\) is the horizontal resolution of the LiDAR, \(\beta=3\) is a multiplication factor, and \(r_p\) is the radial distance to the respective point, as optimized in \cite{pointremove}. As a result, the search radius increases as a function of distance, compensating for the increasing sparsity of lidar point clouds with distance due to the inherent scanning principle of automotive lidars. We classify points with fewer than three neighborhood points as clutter points. If more than 8\% of points are considered clutter, the precipitation is classified as heavy; otherwise, it is classified as light intensity. This value is determined through dedicated experiments from the weather chamber captured in the DENSE dataset. Here, different scenes with varying rain intensities are captured, and we find that 8\% represents a good threshold for many different scenes. The same threshold is used for snow, as a visual qualitative inspection reveals that it translates well to different scenes affected by snow. Fig.~\ref{fig:dror} illustrates an example where all clutter points stemming from heavy snow have been removed. For this example, our method identifies 317 clutter points, amounting to 12\% of the complete point cloud, rightfully labeled as heavy snow. 

\subsection{Classification Method}
\label{sec:classiffication}
We build our network to perform multi-class classification, from a single RGB frame. Instead of single branches for each task, we opt for a unified backbone architecture, providing a single feature map, used by lightweight modules for the different downstream classification tasks. 
As backbone, we adopt the network architecture from \cite{efficientnet}, using as building blocks the mobile inverted bottleneck convolution (MBConv) units. In such way, we obtain a feature map that is shared by six final units (FCBlock), each composed of two fully connected layers and predicting one of the six classes. \newline 
As training objective, instead of the multi-class cross-entropy we employ a multi-class Focal Loss \cite{focalloss}, to ameliorate the problems arising from class unbalance and hard samples. In practice, for each environment condition $c\in\{1,..,6\}$ we compute the Focal loss $\mathcal{L}_c$, leading to the total loss $\mathcal{L}_T$: 
\begin{equation}
    \mathcal{L}_T =  \sum_{c = 1}^{N=6} \alpha_c \mathcal{L}_c = - \sum_{c = 1}^{6} \alpha_c ( 1- p_{ct})^\gamma log(p_{ct})
\end{equation}
Here, $-( 1- p_t)^\gamma$ is the modulating factor used to penalize misclassified samples, adjusted by a scaling hyper-parameter $\gamma$. Each class is also differently weighted using $\alpha_c$.

\subsection{Evaluation Metrics}
\label{sec:eval_metrics}
The evaluation of our RECNet method hinges on five key metrics to provide a comprehensive assessment. Following \cite{weathernet,weatherclassification1}, we evaluate the classification results using the metrics Accuracy, Precision, Recall, F1-Score, and AUPRC (Area Under the Precision-Recall Curve). Accuracy provides a broad perspective on correct predictions, Precision highlights accurate positive predictions, and Recall (Sensitivity) assesses the model's capability to capture all relevant instances, particularly crucial for minimizing false negatives. F1-Score strikes a balance between precision and recall, while AUPRC offers nuanced insights into precision-recall trade-offs, especially in imbalanced scenarios. Collectively, these metrics contribute to a comprehensive evaluation of RECNet's performance across diverse conditions.

\section{Results}
This section delves into a thorough evaluation of our model, featuring extensive ablation studies, quantitative metrics, qualitative results, and visual explanations for predictions. 
\subsection{Ablation Studies}
In this section, we conduct an ablation study to systematically assess the proposed hierarchical labeling strategy and justify the selection of the RECNet architecture and its input.
\paragraph{Labels}
Tab.~\ref{tab:ablation_dataset} shows the results of the proposed weather network trained with different labels. For simplicity, we average Accuracy, F1-Score, and AUPRC over all classification categories. We start with the labels proposed by \cite{SeeThrough}. Training with these labels enables our network to achieve an accuracy of 80\%. Manual relabeling of the data and the utilization of our semi-autolabeling pipeline provide a significant boost to overall performance metrics to 90\% in F1-score. Finally, the incorporation of the proposed hierarchical labeling strategy allows RECNet to improve accuracy by an additional 2\%, reaching a total of 92\%.
\paragraph{Network Architecture and Input}
To validate the choice of the network architecture, we perform an ablation study with different backbones, presented in Tab.~\ref{tab:ablation_backbone}. Specifically, we compare ResNet-50 \cite{resnet}, ViT \cite{transformergoogle}, and EfficientNet-B2 \cite{efficientnet}. With ResNet-50, we achieve an accuracy of 83\%, while ViT allows for an accuracy of up to 88\%. By using EfficientNet-B2, we attain the best performance, achieving an accuracy of 90\%, and motivating our choice of this model as backbone. Moreover, thanks to this backbone architecture, the classification method is lightweight, contributing to high computational efficiency and real-time performance capabilities with 49ms prediction time, essential for autonomous vehicle applications.

\subsection{Classification Results}
\begin{figure*}[!t]
    \centering
    \vspace{2mm}
    \includegraphics[width=1\textwidth]{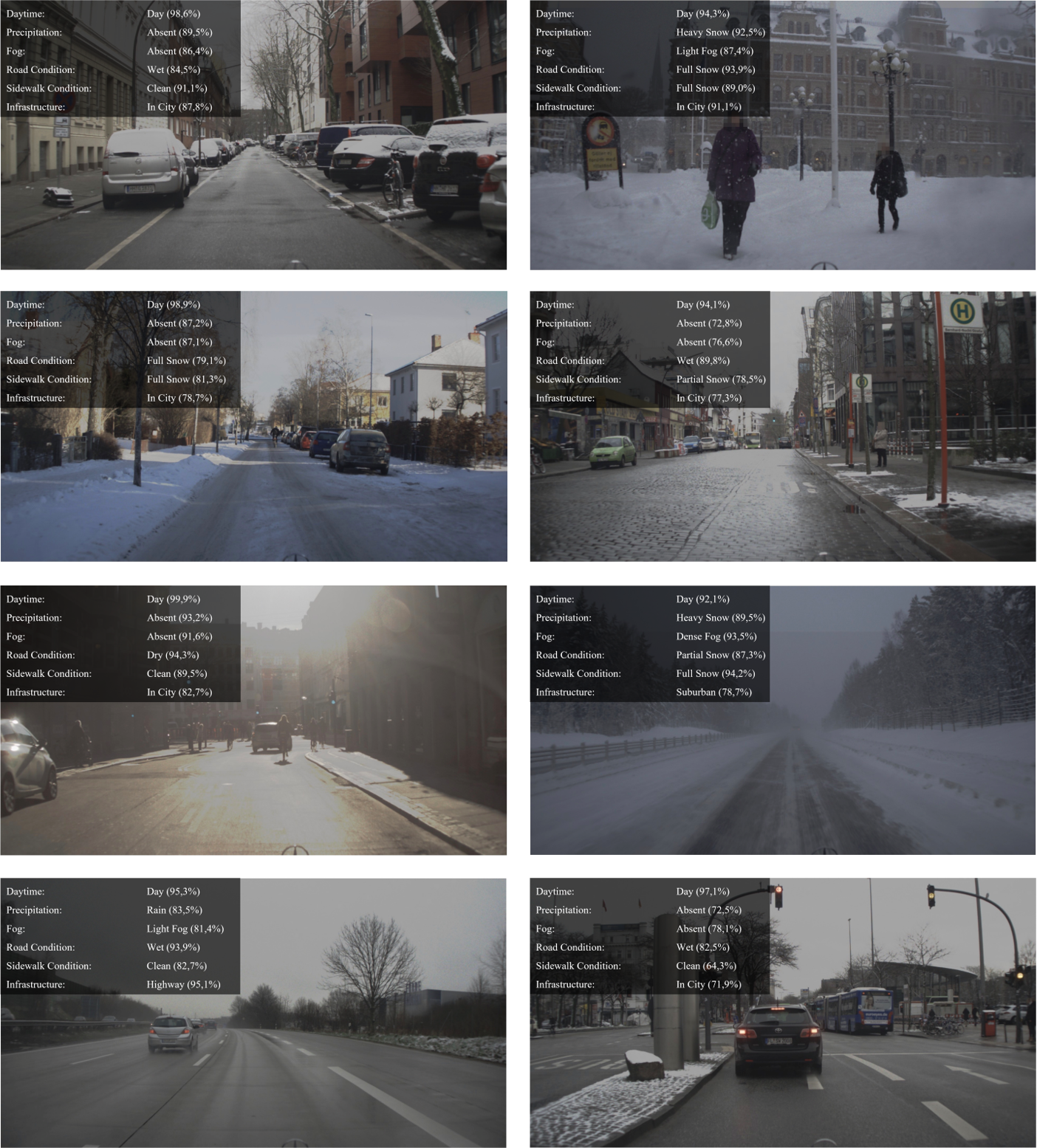}
    \caption{Qualitative classification results of the proposed RECNet approach. Our method is able to predict accurate labels for environment and weather even in challenging conditions.}
    \label{fig:results}
    \vspace{-2mm}
\end{figure*}

In this section, we provide a comprehensive overview of the final classification results for weather prediction, encompassing both quantitative metrics in Tab.~\ref{tab:results} and qualitative assessments in Fig.~\ref{fig:results}.

\paragraph{Quantitative Evaluation}
Tab.~\ref{tab:results} presents a detailed breakdown of the quantitative metrics obtained from our weather classification model. These metrics include accuracy, precision, recall, F1 score, and AUPRC, offering a robust evaluation of the model's proficiency in distinguishing between different weather classes. We employ our novel hierarchical labeling strategy and evaluate the categories "Daytime", "Precipitation", "Fog",  "Road Condition", "Roadside Condition", and "Infrastructure". Our method excels across diverse conditions, achieving an impressive overall classification accuracy exceeding 84\%. Specifically, it attains accuracies of 84\% for "Road Condition", 85\% for "Roadside Condition", 89\% for "Precipitation", 92\% for "Infrastructure", 93\% for "Fog", and 99\% for "Daytime", demonstrating its effectiveness in accurately classifying a wide range of environmental factors. 
The reduced accuracy in classifying road conditions stems from the inherent challenges in discerning subtle visual distinctions between similar states, such as "wet", "partial snow", and "full snow", using RGB images.

\paragraph{Qualitative Evaluation}
In Fig.~\ref{fig:results}, we present qualitative results showcasing RECNet's remarkable performance across various environmental and weather conditions. The visual representations underscore RECNet's ability to provide accurate predictions, capturing nuanced details in diverse scenarios. These results not only highlight the model's robustness but also affirm its efficacy in delivering precise predictions for a broad spectrum of environmental and weather conditions, reinforcing its utility in real-world applications.

\subsection{Visual Explanations for Class Predictions}
\begin{figure}[!t]
    \centering
    \vspace{1mm}
    \begin{subfigure}{0.95\linewidth}
        \centering
        \includegraphics[width=\linewidth]{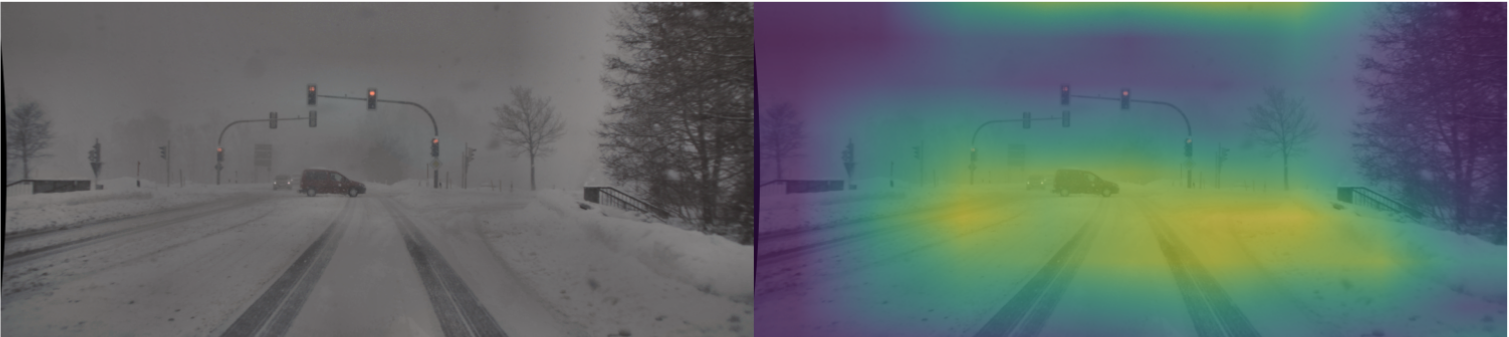}
        \caption{Heat-map for "Road Condition" category}
        \label{fig:gradcam_road}
    \end{subfigure}
    \begin{subfigure}{0.95\linewidth}
        \centering
        \includegraphics[width=\linewidth]{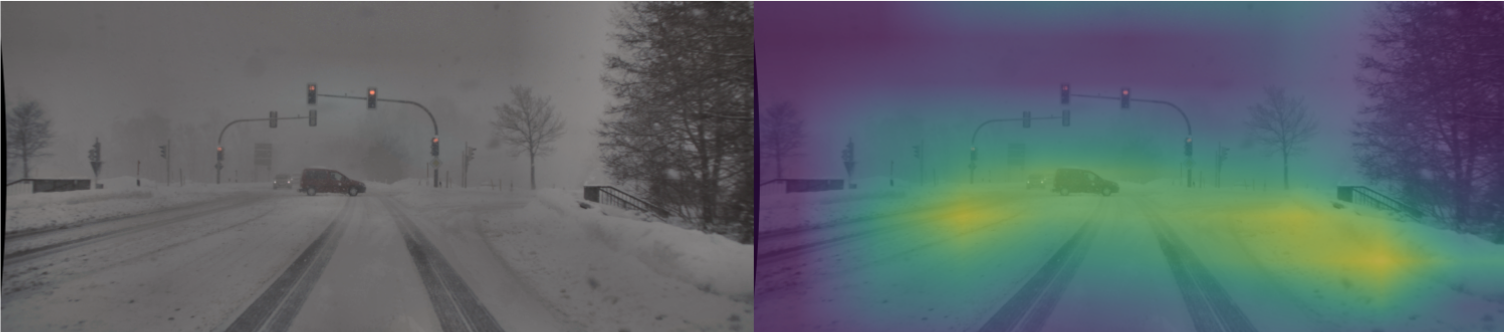}
        \caption{Heat-map for "Sidewalk Condition" category}
        \label{fig:gradcam_side}
    \end{subfigure}
    \caption{Heat-Maps for predictions. We employ Grad-Cam \cite{selvaraju2017grad} to generate visualizations highlighting regions in images that are crucial for network predictions.}
    \label{fig:gradcam}
\end{figure}

In order to understand the predictions of our proposed method better, we employ Grad-Cam (Gradient-weighted Class Activation Mapping) \cite{selvaraju2017grad}. Grad-Cam calculates the gradients of the final convolutional layer and generates heat-maps to highlight relevant regions in the images important for the network's decision. Fig.~\ref{fig:gradcam} shows an example of two heat-maps generated for the categories "Road Conditions" and "Roadside Conditions". It is evident that for the prediction of the "Road Condition" category (Fig.~\ref{fig:gradcam_road}), the focus of the neural network predominantly lies on the road, whereas for "Sidewalk Condition" (Fig.~\ref{fig:gradcam_side}, the neural network's attention is primarily oriented towards the image's periphery, where the sidewalk is located.

\begin{table}[!t]
    \footnotesize
    \centering
    \vspace{2mm}
    \resizebox{.99\linewidth}{!}{
    \begin{tabular}{l|cccc}
            \toprule
            Backbone & Accuracy & F1-Score & AUPRC \\
		\midrule
    ResNet-50~\cite{resnet} & 0.83 & 0.82 & 0.89 \\
    ViT~\cite{transformergoogle} & 0.88 & 0.88 & 0.92 \\
    EfficientNet~\cite{efficientnet} & \textbf{0.92} & \textbf{0.92} & \textbf{0.96} \\
    \bottomrule
    \end{tabular}
    }
    \caption{Ablation of different backbone architectures. We find that EfficientNet outperforms all other baselines, including an EfficientNet with 5 concatenated temporal frames - denoted as temporal - as input. The experiments are carried out on the reannotated dataset.}
    \label{tab:ablation_backbone}
    \vspace{-1mm}
\end{table}

\begin{table}[!t]
    \footnotesize
    \centering
    \vspace{2mm}
    \resizebox{.99\linewidth}{!}{
    \begin{tabular}{l|ccc}
            \toprule
            Dataset & Accuracy & F1-Score & AUPRC \\
		\midrule
  DENSE \cite{SeeThrough} & 0.80 & 0.79 & 0.84 \\
  DENSE (relabeled) & 0.90 & 0.90 & 0.94 \\
  DENSE++ & \textbf{0.92} & \textbf{0.92} & \textbf{0.96} \\
    \bottomrule
    \end{tabular}
    }
    \caption{Dataset ablation. DENSE is the original dataset from \cite{SeeThrough}. DENSE (relabeled) denotes the relabeled DENSE dataset using the original label hierarchy. DENSE++ denotes the relabeled DENSE dataset using the introduced label hierarchy from Sec.~\ref{sec:dataset}. The introduced label hierarchy leads to a significant performance boost in environment condition classification.}
    \label{tab:ablation_dataset}
    \vspace{-1mm}
\end{table}

\begin{table}[!t]
    \footnotesize
    	\setlength{\tabcolsep}{3pt}

    \centering
    \resizebox{.99\linewidth}{!}{
    \begin{tabular}{l|ccccc}
            \toprule
            Category & Accuracy & Precision & Recall & F1-Score & AUPRC \\
		\midrule
  Daytime & 0.99 & 0.99 & 0.99 & 0.99 & 0.99 \\
  Precipitation & 0.89 & 0.88  & 0.89 & 0.88 & 0.89 \\
  Fog & 0.93 & 0.93 & 0.93 & 0.93 & 0.97 \\
  Road Condition & 0.84 & 0.84 & 0.84 & 0.84 & 0.93 \\
  Infrastructure & 0.92 & 0.92 & 0.92 & 0.92 & 0.95 \\
    \bottomrule
    \end{tabular}
    }
    \caption{Classification results. Our method achieves accurate classification results over all predicted environment condition classes considering different metrics.}\label{tab:results}
\end{table}

\section{Conclusion}

We propose RECNet, a unified method for predicting environmental conditions ranging from weather and road conditions to scene settings. To this end, we present DENSE++, an extension of the DENSE dataset, by introducing a new label hierarchy for environmental conditions and subsequently annotating all frames using the semi-automated re-annotation pipeline. The new label hierarchy, allows to achieve state-of-the-art classification results, while the chosen classification architecture enables real-time capability in the context of autonomous driving. Future work will explore the multi-modal and temporal data provided within the DENSE dataset, allowing for more complex weather identification networks that incorporate disturbance models across different modalities, cross-sensor redundancies, and broken temporal consistencies due to falling rain or snow particles. 

\section{Acknowledgments}
The research leading to these results is part of the AI-SEE project, which is a co-labelled PENTA and EURIPIDES2
project endorsed by EUREKA. Co-funding is provided by
the following national funding authorities: Austrian Research
Promotion Agency (FFG), Business Finland, Federal Ministry
of Education and Research (BMBF) and National Research
Council of Canada Industrial Research Assistance Program
(NRC-IRAP).

\bibliographystyle{IEEEtran}
\bibliography{bib.bib}

\end{document}